\documentclass[runningheads]{llncs}
\usepackage[T1]{fontenc}
\usepackage{graphicx}
\usepackage{multirow}
\usepackage{color}
\begin{document}
\title{KSE-Web: An Analysis of Hybrid Retrieval and LLM-Assisted Query Expansion for Low-Resource Khmer Semantic Search}
\titlerunning{Khmer Semantic Search: Hybrid Retrieval and LLM-Assisted Query}
%
\author{Nimol Thuon}
\authorrunning{Nimol Thuon}
%
%
\maketitle              
\begin{abstract}
As a low-resource language, Khmer presents several retrieval challenges, including limited annotated data, ambiguous word boundaries, weak support in multilingual embedding models, and frequent mixed Khmer--English usage. This paper presents \textbf{KSE-Web}, an analysis of hybrid retrieval and LLM-assisted query expansion for Khmer semantic search. We construct the dataset from approximately 17K candidate Khmer titles and retain 3K cleaned full-text Khmer documents after filtering, normalization, deduplication, and document-length control. The dataset includes 300 manually reviewed user-style Khmer search queries and silver relevance labels with partial human verification. We evaluate character n-gram BM25, multilingual dense retrieval, hybrid BM25+dense retrieval, and LLM-assisted query expansion using Qwen2.5 models. Experimental results show that BM25 achieves the strongest overall performance, reaching 0.943 Recall@10 and 0.876 nDCG@10. Hybrid BM25+dense retrieval performs comparably, achieving 0.929 Recall@10 and 0.871 nDCG@10, while dense retrieval alone performs lower. LLM-assisted query expansion does not outperform non-expanded retrieval; however, Qwen2.5-3B produces substantially stronger expanded-query results than Qwen2.5-0.5B, suggesting that LLM size and expansion quality matter for low-resource Khmer retrieval. Our analysis further shows that direct LLM expansion can introduce topic drift, generic terms, and noisy reformulations, while simple filtering may remove useful semantic cues. These findings highlight both the potential and limitations of LLM-assisted retrieval for Khmer semantic search and provide a foundation for future Khmer retrieval datasets with stronger human-verified annotations and Khmer-aware retrieval models. The dataset and documentation will be made available at \url{github.com/back-kh/Khmer-Semantic-Search}.

\end{abstract}
\section{Introduction}
\label{sec:introduction}

Large language models (LLMs) have recently achieved strong performance across natural language processing, computer vision, and document understanding tasks. Their ability to generate paraphrases, reformulate questions, and provide contextual expansions has motivated their use in retrieval-oriented applications, including semantic search and retrieval-augmented systems \cite{yang2024qwen25,wang2023query2doc,chang2024survey}. However, the effectiveness of LLM-assisted retrieval remains uneven across languages. In particular, low-resource languages often lack large-scale retrieval datasets, high-quality relevance judgments, robust tokenization tools, and strong pretrained representations. As a result, methods that perform well for English or other high-resource languages may not directly transfer to languages with limited digital resources \cite{abe2025llm,shen2024language,sindhujan2025llms}.
\begin{figure}[t] 
\centering \includegraphics[width=\textwidth]{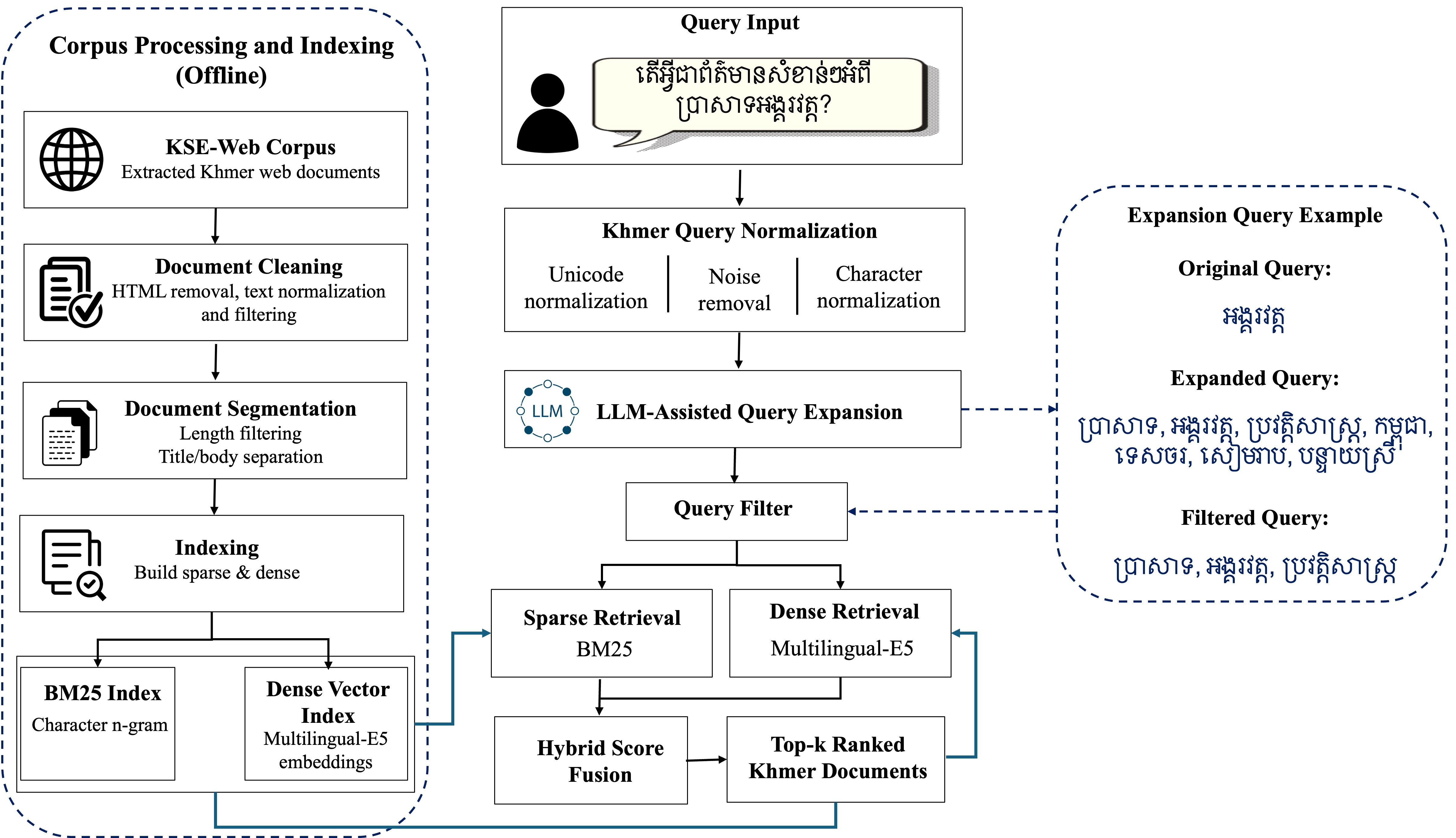} 
\caption{Overview of the KSE-Web retrieval framework. The offline stage constructs sparse and dense indexes from cleaned Khmer web documents. At inference time, a Khmer query is normalized and optionally expanded using an LLM. The original or expanded query is then used for sparse BM25 retrieval, dense retrieval with Multilingual-E5, and hybrid score fusion to produce top-$k$ ranked Khmer documents.} 
\label{fig:kse_framework} 
\end{figure}
Khmer is one such language. Although it is widely used in Cambodia, Khmer remains underexplored in information retrieval and semantic search research in the modern era. Khmer web retrieval presents several practical challenges ~\cite{thuon2022syllable,thuon2024}. First, existing Khmer semantic search work has mainly focused on domain-specific retrieval and conventional preprocessing, while limited attention has been given to hybrid retrieval strategies or LLM-assisted semantic search models~\cite{thuon2024khmer,thuon2024ksw}. Second, Khmer does not use whitespace as a reliable word boundary, making standard word-level tokenization and retrieval less straightforward. Third, real-world Khmer web queries often include informal expressions, spelling variations, named entities, and mixed Khmer--English usage. Finally, multilingual dense retrieval models may not fully capture Khmer semantic similarity because Khmer is underrepresented in many multilingual training and evaluation resources. These challenges make Khmer a valuable case for studying retrieval in low-resource language settings \cite{buoy2021khmer,kaing2025prahokbart}.

This paper presents \textbf{KSE-Web}, an analysis of hybrid retrieval and LLM-assisted query expansion for low-resource Khmer semantic search using a Khmer web retrieval dataset. The overall framework is shown in Fig.~\ref{fig:kse_framework}. KSE-Web separates offline corpus processing from online query processing. In the offline stage, cleaned Khmer web documents are used to construct both sparse and dense indexes. In the online stage, a user query is normalized and optionally expanded using an LLM before being passed to sparse retrieval, dense retrieval, and hybrid score fusion.

To support this analysis, we construct a Khmer web retrieval dataset from approximately 17K candidate Khmer web titles and retain 3K cleaned full-text Khmer documents after filtering, normalization, deduplication, and document-length control. The dataset construction pipeline is summarized in Fig.~\ref{fig:dataset_pipeline}. The dataset includes 300 manually reviewed user-style Khmer queries and 5,412 query--document relevance labels generated through silver labeling with partial human verification. The queries cover multiple search styles, including short keyword queries, question-style queries, informal user-style queries, and mixed Khmer--English queries.

We evaluate character n-gram BM25, multilingual dense retrieval, hybrid BM25+dense retrieval, and LLM-assisted query expansion using Qwen2.5 models. Our results show that character n-gram BM25 remains the strongest overall method, achieving 0.943 Recall@10 and 0.876 nDCG@10. Hybrid retrieval performs comparably, reaching 0.929 Recall@10 and 0.871 nDCG@10, while dense retrieval alone performs substantially lower. The main performance trends are summarized in Fig.~\ref{fig:retrieval_results}. LLM-assisted query expansion does not outperform non-expanded retrieval, but Qwen2.5-3B produces substantially stronger expanded-query results than Qwen2.5-0.5B. Qualitative examples in Fig.~\ref{fig:llm_behavior} further show that LLM expansion may introduce topic drift, generic terms, and noisy reformulations, while simple filtering can remove useful semantic cues together with noise.

The main contributions of this work are as follows:
\begin{itemize}
    \item We construct a Khmer web retrieval dataset containing 3K cleaned full-text Khmer web documents, 300 manually reviewed user-style queries, and 5,412 silver relevance labels with partial human verification.
    \item We present KSE-Web, a unified analysis framework for low-resource Khmer semantic search covering sparse, dense, hybrid, and LLM-assisted retrieval settings.
    \item We show that character n-gram BM25 is a strong and necessary baseline for Khmer web retrieval, while hybrid BM25+dense retrieval performs closely but does not surpass BM25.
    \item We analyze LLM-assisted query expansion using Qwen2.5 models and show that larger LLMs produce more useful expansions than smaller LLMs, but direct query expansion still remains below non-expanded retrieval.
    \item We provide qualitative analysis of LLM expansion behavior, including helpful expansion, topic drift, over-expansion, generic-term insertion, and filtering effects.
\end{itemize}

\section{Related Work}
\label{sec:related_work}

\subsection{Low-Resource and Multilingual Information Retrieval}

Information retrieval for low-resource languages remains challenging because many languages lack large-scale corpora, standardized query sets, relevance judgments, and language-specific preprocessing tools. Recent benchmarks such as BEIR, Mr. TyDi, mMARCO, and MIRACL have advanced retrieval evaluation across diverse domains and languages~\cite{thakur2021beir,zhang2021mr,bonifacio2021mmarco,zhang2023miracl}. However, Khmer remains much less represented in retrieval research compared with high-resource and better-studied multilingual settings. This motivates the need for Khmer-specific retrieval datasets and evaluation protocols.

Khmer retrieval presents both linguistic and practical challenges~\cite{buoy2021khmer,thuon2024khmer}. Khmer does not use whitespace as a reliable word boundary, making word-level tokenization difficult. In addition, real-world Khmer web content often includes mixed Khmer--English terms, transliterated names, named entities, informal expressions, and spelling variations. These issues affect both sparse retrieval methods, which depend on lexical overlap, and dense retrieval methods, which depend on learned semantic representations~\cite{robertson2009probabilistic,karpukhin2020dense}. Recent hybrid and LLM-assisted retrieval methods offer promising directions for low-resource retrieval~\cite{wang2023query2doc,coleman2026comparing,jian2024large,takehi2025llm}. In this work, we focus on Khmer web retrieval and construct a dataset for controlled analysis of sparse, dense, hybrid, and LLM-assisted retrieval methods.

\subsection{Sparse, Dense, and Hybrid Retrieval}

Sparse lexical retrieval remains a strong baseline in information retrieval. BM25 is widely used because it is efficient, interpretable, and effective for ranking documents using lexical evidence~\cite{robertson2009probabilistic}. For languages such as Khmer, however, word-level BM25 is affected by tokenization difficulty. Character n-gram BM25 offers a practical alternative because it does not require explicit word segmentation and can capture partial lexical overlap between queries and documents. This makes character-level sparse retrieval suitable for Khmer web search.

Dense retrieval represents queries and documents as continuous vectors and ranks documents by vector similarity. Dense Passage Retrieval (DPR) showed the effectiveness of dual-encoder retrieval for open-domain question answering~\cite{karpukhin2020dense}. More recent embedding models, such as E5, provide general-purpose text representations for retrieval, clustering, and classification~\cite{wang2022text}. Multilingual embedding models are useful for low-resource languages because they can be applied without training a language-specific retriever. However, their performance depends on how well the target language is represented during training. For Khmer, off-the-shelf multilingual embeddings may still struggle with semantic similarity, named entities, and mixed-script usage.

Hybrid retrieval combines sparse and dense retrieval signals. Sparse retrieval is effective for exact lexical matching, named entities, and rare terms, while dense retrieval can capture paraphrases and semantically related expressions. In this paper, we use a simple score-fusion strategy between character n-gram BM25 and multilingual-E5 dense retrieval. This allows us to examine whether dense retrieval adds useful semantic signals beyond character-level lexical matching for Khmer search.

\subsection{LLM-Assisted Query Expansion and Relevance Labeling}

Query expansion improves retrieval by reformulating a query or adding related terms to better match relevant documents. Traditional methods include relevance feedback, pseudo-relevance feedback, and term selection~\cite{carpineto2012survey,chang2024survey}. Recent studies have also explored LLM-based query expansion. Query2doc generates pseudo-documents to expand queries for sparse and dense retrieval~\cite{wang2023query2doc}, while HyDE generates hypothetical documents for zero-shot dense retrieval without relevance labels~\cite{gao2023hyde}. These methods suggest that generated text can provide useful retrieval context, especially for short or ambiguous queries.

However, LLM-assisted expansion can introduce errors. Generated text may broaden the original intent, add generic terms, or include related but non-discriminative words. These risks are more important for low-resource languages, where LLMs may have weaker language-specific knowledge and generated terms may not match the target collection~\cite{shen2024language,sindhujan2025llms}. In this work, we analyze Qwen2.5-based query expansion for Khmer retrieval~\cite{yang2024qwen25} by comparing Qwen2.5-0.5B and Qwen2.5-3B. The larger model produces more useful expansions than the smaller model, but direct expansion still does not outperform strong non-expanded retrieval baselines.

Reliable relevance labels are necessary for retrieval evaluation, but full manual annotation is costly, especially for low-resource languages. Many retrieval datasets therefore use pooling, where candidate documents are retrieved by one or more systems and then judged for relevance. In resource-constrained settings, silver labels provide a practical starting point, although they may introduce bias toward the systems used for pooling. In KSE-Web, we construct silver relevance labels using source-document matching, BM25-based candidate retrieval, rule-based relevance assignment, and partial human verification. Therefore, we view the experiments as a controlled analysis of Khmer web retrieval behavior rather than a final gold-standard benchmark.

\begin{figure}[t] 
\centering \includegraphics[width=\textwidth]{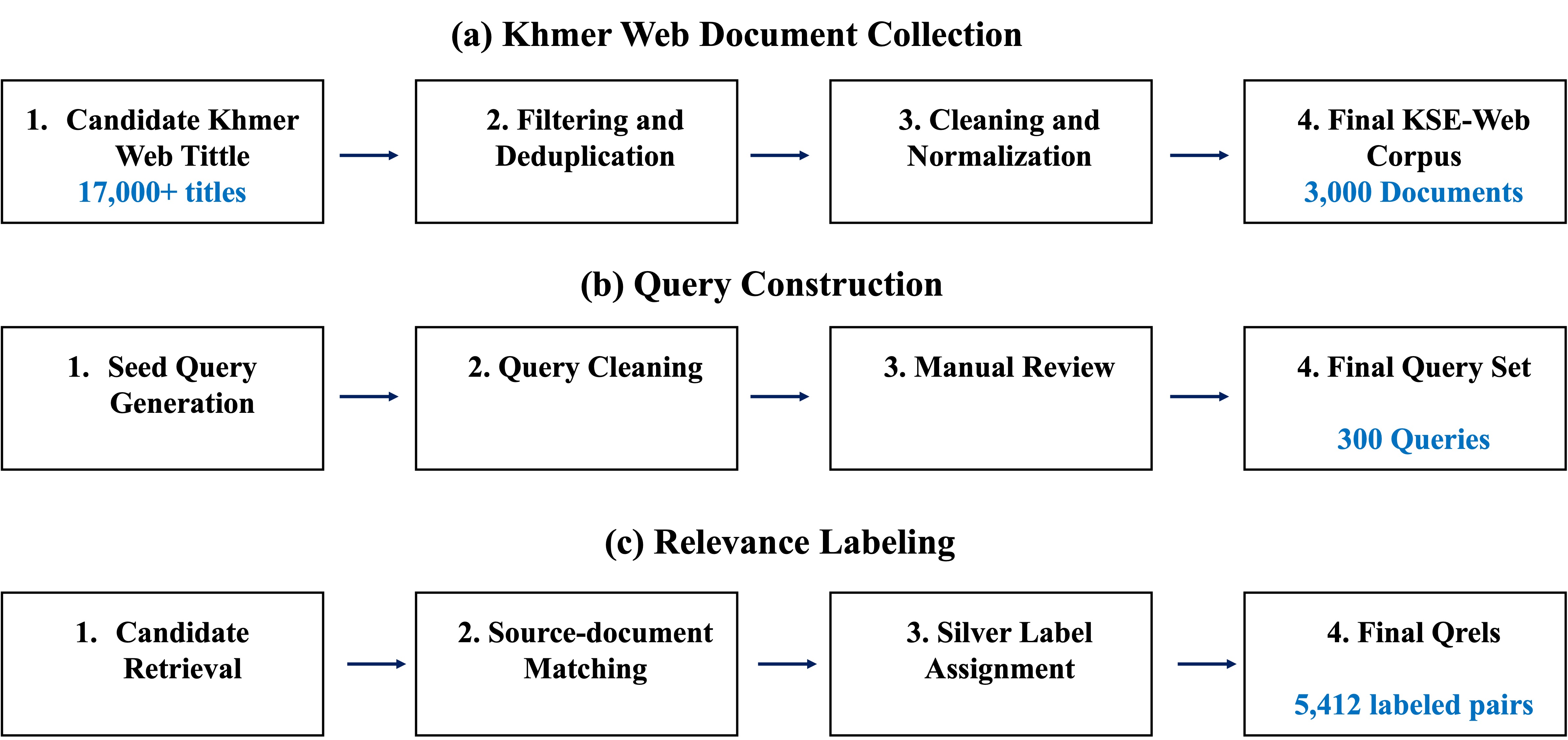} 
\caption{Construction pipeline of the KSE-Web retrieval dataset. The dataset is built from approximately 17K candidate Khmer web titles, filtered and cleaned into 3K full-text Khmer web documents, and paired with 300 manually reviewed user-style queries. Candidate retrieval, source-document matching, silver relevance labeling, and partial human verification produce 5,412 query--document relevance labels.} \label{fig:dataset_pipeline} 
\end{figure}

\section{KSE-Web Dataset Construction}
\label{sec:dataset}

This section describes the construction of KSE-Web, a Khmer web retrieval dataset designed for controlled analysis of sparse, dense, hybrid, and LLM-assisted retrieval methods. Since publicly available Khmer retrieval datasets with user-style queries and relevance labels are limited, we construct a new resource consisting of cleaned Khmer web documents, manually reviewed queries, and silver relevance labels with partial human verification. The overall pipeline is shown in Fig.~\ref{fig:dataset_pipeline}.

\subsection{Document Collection, Categories, and Preprocessing}
\label{subsec:document_collection}

We construct the document collection from publicly available Khmer web content. Starting from approximately 17K candidate Khmer web titles, we collect the corresponding pages and process them into a clean full-text corpus. Noisy, duplicated, very short, and non-document-like pages are removed through title filtering, Khmer-content filtering, duplicate removal, and basic quality control. The extracted text is further cleaned using HTML removal, Unicode normalization, whitespace normalization, and noise filtering. Long documents are truncated after cleaning to keep document-level retrieval efficient and comparable.

After filtering and cleaning, the final KSE-Web corpus contains 3K full-text Khmer web documents. Each document includes a document identifier, title, cleaned body text, source URL, category metadata, and document-length information. Each document is assigned to one of four broad categories: \textit{public service}, \textit{education}, \textit{tourism and culture}, and \textit{general news/information}. These categories reflect common Khmer web search needs and support domain-level retrieval analysis. Category assignment is based on rule-based keywords over titles and cleaned text, followed by partial manual inspection to reduce obvious errors.

\subsection{Query Construction and Silver Relevance Labeling}
\label{subsec:query_labeling}

We create 300 user-style Khmer search queries for document-level retrieval evaluation. Candidate queries are generated from source document titles using template-based transformations, allowing each query to be linked to at least one relevant source document. The query set covers four types: \textit{short}, \textit{question-style}, \textit{informal}, and \textit{mixed Khmer--English}. To improve quality, all queries are manually reviewed by a Khmer speaker to remove duplicated, unnatural, overly broad, or noisy template-generated queries.

Fully manual relevance annotation is costly in low-resource settings, so KSE-Web uses silver relevance labels with partial human verification. For each query, we retrieve the top-20 candidate documents using character n-gram BM25. The source document is added to the candidate pool when it is not retrieved in the top-20 results. Each query--document pair is assigned a three-level relevance label: 2 for highly relevant, 1 for partially relevant, and 0 for non-relevant. Label assignment uses source-document matching, title-query overlap, BM25 ranking, category agreement, rule-based scoring, and partial human inspection.

\begin{table}[t]
\centering
\caption{Statistics of the KSE-Web dataset.}
\label{tab:dataset_statistics}
\begin{tabular}{ll}
\hline
Component & Count \\
\hline
Candidate Khmer web titles & $\sim$17K \\
Khmer web documents & 3000 \\
Document categories & 4 \\
User-style queries & 300 \\
Labeled query-document pairs & 5412 \\
Relevance scale & 0, 1, 2 \\
\hline
\end{tabular}
\end{table}

\subsection{Dataset Statistics and Scope}
\label{subsec:dataset_statistics}

Table~\ref{tab:dataset_statistics} summarizes the main statistics of KSE-Web. The dataset contains 3K cleaned Khmer web documents, 300 manually reviewed queries, and 5,412 labeled query--document pairs. The three-level relevance scale supports both binary relevance evaluation and graded ranking metrics such as nDCG.

KSE-Web is intended as a practical Khmer web retrieval dataset rather than a final large-scale gold-standard benchmark. The current version focuses on web-extracted Khmer text and does not include PDFs, scanned pages, or OCR-based documents. In addition, some queries are derived from document titles, and the silver labels partly rely on BM25-based candidate retrieval, which may introduce lexical bias. These limitations are considered in the experimental analysis and motivate future work on larger human-verified Khmer retrieval datasets.

\section{Overall Retrieval Framework}
\label{sec:framework}

This section describes the overall retrieval framework used to evaluate low-resource Khmer semantic search in KSE-Web. Given a Khmer user query, the goal is to retrieve a ranked list of relevant documents from the KSE-Web document collection. As shown in Fig.~\ref{fig:kse_framework}, the framework contains two main stages: an offline indexing stage and an online retrieval stage. The offline stage prepares sparse and dense indexes from cleaned Khmer web documents, while the online stage processes user queries through optional LLM-assisted expansion, sparse retrieval, dense retrieval, and hybrid score fusion.

\subsection{Task Definition}
\label{subsec:task_definition}

Let $\mathcal{D} = \{d_1, d_2, \ldots, d_N\}$ denote the KSE-Web document collection, where each document $d_i$ contains a title and cleaned Khmer web text. Given a user query $q$, the retrieval task is to produce a ranked list of documents:
\[
R(q) = [d_{(1)}, d_{(2)}, \ldots, d_{(k)}],
\]
where documents appearing earlier in the list are expected to be more relevant to the query. The task is evaluated at the document level rather than the passage level. Each method retrieves documents from the same KSE-Web corpus and is evaluated using the same silver relevance labels described in Section~\ref{sec:dataset}. Relevance is represented using a three-level scale: 2 for highly relevant, 1 for partially relevant, and 0 for non-relevant.

\subsection{Sparse Retrieval with Character n-gram BM25}
\label{subsec:sparse_retrieval}

As the sparse lexical baseline, we use BM25 over character n-gram representations. BM25 is a strong retrieval method for lexical matching, but word-level BM25 requires reliable tokenization. This is difficult for Khmer because whitespace does not consistently indicate word boundaries. To avoid reliance on Khmer word segmentation, both queries and documents are represented using overlapping character n-grams.

For each query and document, we remove unnecessary whitespace and extract character n-grams with $n \in \{2,3,4\}$. The document representation is created from the concatenation of the document title and cleaned body text. The sparse retrieval score is computed as:
\[
S_{\mathrm{BM25}}(q,d) = \mathrm{BM25}(G(q), G(d)),
\]
where $G(\cdot)$ denotes the character n-gram representation. This design allows the sparse retriever to capture partial lexical overlap, named entities, spelling variants, and short Khmer phrases without requiring explicit word segmentation.

\subsection{Dense Retrieval with Multilingual Embeddings}
\label{subsec:dense_retrieval}

For dense retrieval, we use a multilingual sentence embedding model to encode queries and documents into vector representations. In our experiments, we use multilingual-E5-small as the dense retriever. Following the E5 input format, queries are prefixed with \texttt{query:}, while documents are prefixed with \texttt{passage:}. Each document embedding is computed from the concatenation of its title and cleaned text.

Let $\mathbf{e}_q$ denote the embedding of query $q$, and let $\mathbf{e}_d$ denote the embedding of document $d$. Dense retrieval ranks documents using cosine similarity:
\[
S_{\mathrm{dense}}(q,d) =
\cos(\mathbf{e}_q, \mathbf{e}_d).
\]
Dense retrieval is intended to capture semantic similarity beyond exact lexical overlap. However, because Khmer is less represented in many multilingual training resources, the effectiveness of off-the-shelf multilingual embeddings for Khmer retrieval remains uncertain. This motivates our comparison between dense retrieval, sparse retrieval, and hybrid retrieval.

\subsection{Hybrid Score Fusion}
\label{subsec:hybrid_retrieval}

Hybrid retrieval combines sparse and dense retrieval signals. Sparse retrieval is effective for exact lexical matching, named entities, and rare terms, while dense retrieval can capture broader semantic relatedness. To combine these two signals, we first normalize BM25 and dense scores for each query using min--max normalization:
\[
\hat{S}(q,d) =
\frac{S(q,d) - \min_{d' \in \mathcal{D}} S(q,d')}
{\max_{d' \in \mathcal{D}} S(q,d') - \min_{d' \in \mathcal{D}} S(q,d')}.
\]
The final hybrid score is then computed as:
\[
S_{\mathrm{hybrid}}(q,d) =
\alpha \hat{S}_{\mathrm{BM25}}(q,d)
+
(1-\alpha)\hat{S}_{\mathrm{dense}}(q,d),
\]
where $\alpha$ controls the contribution of sparse retrieval. In this study, we set $\alpha = 0.5$ to give equal weight to BM25 and dense retrieval. Documents are ranked according to $S_{\mathrm{hybrid}}(q,d)$.

\subsection{LLM-Assisted Query Expansion}
\label{subsec:llm_expansion}

We analyze whether LLM-assisted query expansion can improve Khmer semantic search. Given an original query $q$, an instruction-tuned LLM generates an expanded query $q^{+}$ by adding related terms, paraphrases, or commonly used Khmer--English expressions. In this work, we evaluate Qwen2.5-0.5B-Instruct and Qwen2.5-3B-Instruct to study the effect of LLM size on expansion quality.

The LLM is prompted to preserve the original query meaning, keep named entities unchanged, and add only retrieval-useful terms. The expanded query is then used as input to BM25, dense retrieval, or hybrid retrieval. Formally, retrieval with an expanded query is computed as:
\[
S(q^{+}, d),
\]
where $S$ may correspond to sparse, dense, or hybrid retrieval.

In addition to raw LLM expansion, we evaluate a filtered expansion variant. The filtering step removes repeated terms, overly generic English words, broad Khmer terms, and excessive expansion tokens. The filtered query is represented as:
\[
q^{f} = q \oplus F(q^{+}),
\]
where $F(\cdot)$ denotes the filtering function and $\oplus$ denotes concatenation with the original query. This variant tests whether simple filtering can reduce expansion noise and topic drift.

\subsection{Evaluated Retrieval Variants}
\label{subsec:retrieval_variants}

Using the components above, we evaluate the following retrieval variants:
\begin{itemize}
    \item \textbf{BM25-char-ngram}: sparse retrieval using Khmer character n-grams.
    \item \textbf{Dense}: dense retrieval using multilingual-E5-small embeddings.
    \item \textbf{Hybrid-BM25-Dense}: score fusion of BM25-char-ngram and dense retrieval.
    \item \textbf{BM25 + LLM Expansion}: sparse retrieval using LLM-expanded queries.
    \item \textbf{Dense + LLM Expansion}: dense retrieval using LLM-expanded queries.
    \item \textbf{Hybrid + LLM Expansion}: hybrid retrieval using LLM-expanded queries.
    \item \textbf{Filtered LLM Expansion}: retrieval using filtered expanded queries.
\end{itemize}

This framework allows us to examine not only which retrieval method performs best, but also how LLM-assisted query expansion interacts with sparse, dense, and hybrid retrieval in low-resource Khmer semantic search.

\section{Experiments}
\label{sec:experiments}

This section describes the experimental setup used to evaluate sparse, dense, hybrid, and LLM-assisted retrieval methods on KSE-Web. The goal is not only to identify the strongest retrieval method, but also to analyze how different retrieval strategies behave in a low-resource Khmer semantic search setting.

\subsection{Dataset and Evaluation Setting}
\label{subsec:experimental_dataset}

We conduct experiments on the KSE-Web retrieval dataset described in Section~\ref{sec:dataset}. The dataset contains 3,000 cleaned full-text Khmer web documents and 300 manually reviewed user-style Khmer queries. The queries are balanced across four broad document categories and cover four query types: short, question-style, informal, and mixed Khmer--English queries. Each retrieval method searches over the same document collection and returns the top-20 ranked documents for each query.

Evaluation is performed using the silver relevance labels described in Section~\ref{sec:dataset}. Each query--document pair is assigned a three-level relevance label: 2 for highly relevant, 1 for partially relevant, and 0 for non-relevant. For Recall, Precision, and MRR, labels 1 and 2 are treated as relevant. For nDCG, the original graded relevance labels are used.

\subsection{Compared Methods}
\label{subsec:compared_methods}

We compare the following retrieval methods.

\paragraph{BM25-char-ngram.}
This is the main sparse retrieval baseline. Queries and documents are represented using Khmer character n-grams with $n \in \{2,3,4\}$, and documents are ranked using BM25. This method is designed to avoid dependence on Khmer word segmentation.

\paragraph{Dense retrieval.}
Dense retrieval uses multilingual-E5-small to encode queries and documents into vector representations. Queries are prefixed with \texttt{query:}, and documents are prefixed with \texttt{passage:}. Documents are ranked by cosine similarity.

\paragraph{Hybrid BM25+dense retrieval.}
Hybrid retrieval combines normalized BM25 and dense retrieval scores using linear interpolation with $\alpha=0.5$. This setting gives equal weight to sparse and dense retrieval signals.

\paragraph{LLM-assisted query expansion.}
We evaluate query expansion using Qwen2.5-0.5B-Instruct and Qwen2.5-3B-Instruct. Given an original Khmer query, the LLM generates an expanded query while being instructed to preserve the original meaning, keep named entities unchanged, and add only retrieval-useful terms. The expanded queries are then used with BM25, dense retrieval, and hybrid retrieval.

\paragraph{Filtered LLM expansion.}
We also evaluate a filtered expansion variant. The filtering step removes repeated terms, generic English words, broad Khmer terms, and excessive expansion tokens. This variant tests whether simple filtering can reduce expansion noise and topic drift.

\subsection{Evaluation Metrics}
\label{subsec:evaluation_metrics}

We evaluate retrieval performance using standard top-$k$ metrics at cutoffs 5 and 10, including Recall@k, Precision@k, MRR@k, and nDCG@k. Recall@k measures how many relevant documents are retrieved, Precision@k measures the relevance ratio among the top-$k$ results, MRR@k measures the rank of the first relevant document, and nDCG@k evaluates ranking quality with graded relevance labels. All metrics are averaged over the full query set.

\subsection{Implementation Details}
\label{subsec:implementation_details}

All experiments are implemented in Python. BM25 retrieval uses character n-gram tokenization over the concatenation of document title and cleaned body text. Dense retrieval uses multilingual-E5-small with normalized embeddings and cosine similarity. For hybrid retrieval, BM25 and dense scores are min--max normalized per query before score fusion. LLM-assisted query expansion is performed using Qwen2.5-0.5B-Instruct and Qwen2.5-3B-Instruct with deterministic decoding. The maximum number of generated tokens is limited to keep the expanded queries short and retrieval-focused. For each method, the top-20 documents are retrieved, and metrics are reported at cutoffs 5 and 10.
\begin{figure}[t] \centering \includegraphics[width=\textwidth]{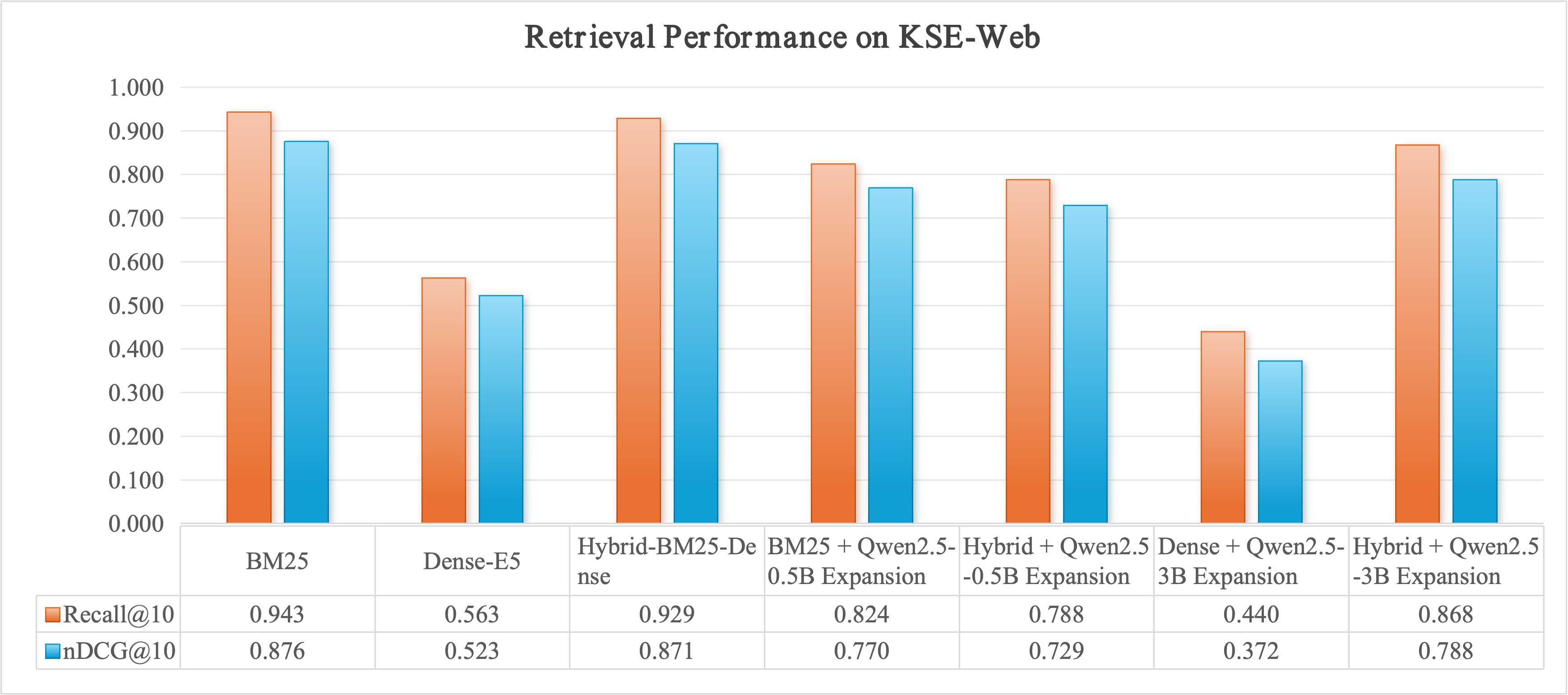} 
\caption{Summary of retrieval performance on KSE-Web. BM25 achieves the strongest overall performance, while hybrid retrieval remains close. LLM-assisted expansion improves with Qwen2.5-3B relative to Qwen2.5-0.5B, but does not surpass non-expanded BM25 or hybrid retrieval. Full quantitative results are reported in Table~\ref{tab:main_results}.} 
\label{fig:retrieval_results} \end{figure}
\section{Results and Analysis}
\label{sec:results}

This section reports the retrieval results on KSE-Web and analyzes the behavior of sparse, dense, hybrid, and LLM-assisted retrieval methods. The main retrieval trends are summarized visually in Fig.~\ref{fig:retrieval_results}, while Table~\ref{tab:main_results} reports the full quantitative results.

\subsection{Overall Retrieval Performance}
\label{subsec:overall_results}

Table~\ref{tab:main_results} shows that character n-gram BM25 achieves the strongest overall performance, reaching 0.943 Recall@10 and 0.876 nDCG@10. This confirms that character-level sparse retrieval is a highly competitive baseline for Khmer web search. Since Khmer does not reliably mark word boundaries with whitespace, character n-grams provide a robust way to capture partial lexical overlap, named entities, and short phrases without relying on explicit word segmentation.

Hybrid BM25+dense retrieval performs very closely to BM25, reaching 0.929 Recall@10 and 0.871 nDCG@10. This suggests that dense representations provide some complementary retrieval signal, but they do not surpass the sparse lexical baseline in the current KSE-Web setting. Dense retrieval alone performs substantially lower, with 0.563 Recall@10 and 0.523 nDCG@10. This indicates that multilingual-E5-small captures some Khmer semantic similarity, but remains limited compared with character n-gram BM25.

\begin{table}[t]
\centering
\caption{Retrieval performance on KSE-Web at cutoffs 5 and 10. BM25 denotes character n-gram BM25, Dense denotes multilingual-E5-small retrieval, and Hybrid denotes BM25+dense score fusion. Qwen-0.5B and Qwen-3B indicate LLM-assisted query expansion using Qwen2.5-0.5B and Qwen2.5-3B, respectively. F denotes filtered expansion. Best results are shown in bold.}

\label{tab:main_results}
\resizebox{\textwidth}{!}{
\begin{tabular}{lcccccccc}
\hline
\textbf{Method} & \textbf{R@5} & \textbf{P@5} & \textbf{MRR@5} & \textbf{nDCG@5} & \textbf{R@10} & \textbf{P@10} & \textbf{MRR@10} & \textbf{nDCG@10} \\
\hline
BM25 & \textbf{0.895} & \textbf{0.291} & 0.905 & \textbf{0.863} & \textbf{0.943} & \textbf{0.166} & 0.906 & \textbf{0.876} \\
Dense & 0.504 & 0.151 & 0.549 & 0.504 & 0.563 & 0.089 & 0.558 & 0.523 \\
Hybrid & 0.873 & 0.279 & \textbf{0.911} & 0.854 & 0.929 & 0.157 & \textbf{0.914} & 0.871 \\
\hline
BM25 + Qwen-0.5B & 0.774 & 0.238 & 0.794 & 0.753 & 0.824 & 0.133 & 0.798 & 0.770 \\
BM25 + Qwen-0.5B-F & 0.761 & 0.229 & 0.768 & 0.730 & 0.810 & 0.128 & 0.772 & 0.747 \\
Dense + Qwen-0.5B  & 0.312 & 0.087 & 0.324 & 0.299 & 0.365 & 0.052 & 0.332 & 0.318 \\
Dense + Qwen-0.5B-F  & 0.319 & 0.087 & 0.306 & 0.291 & 0.361 & 0.051 & 0.312 & 0.304 \\
Hybrid + Qwen-0.5B  & 0.732 & 0.220 & 0.763 & 0.710 & 0.788 & 0.124 & 0.767 & 0.729 \\
Hybrid + Qwen-0.5B-F  & 0.707 & 0.208 & 0.745 & 0.688 & 0.767 & 0.118 & 0.749 & 0.709 \\
\hline
Dense + Qwen-3B  & 0.373 & 0.109 & 0.381 & 0.350 & 0.440 & 0.066 & 0.391 & 0.372 \\
Dense + Qwen-3B-F  & 0.349 & 0.100 & 0.350 & 0.325 & 0.410 & 0.060 & 0.360 & 0.347 \\
Hybrid + Qwen-3B  & 0.798 & 0.245 & 0.806 & 0.766 & 0.868 & 0.143 & 0.810 & 0.788 \\
Hybrid + Qwen-3B-F  & 0.769 & 0.231 & 0.773 & 0.733 & 0.838 & 0.134 & 0.779 & 0.755 \\
\hline
\end{tabular}
}
\end{table}

The gap between BM25 and dense retrieval suggests that off-the-shelf multilingual embeddings are not yet sufficient for Khmer semantic retrieval. This may be due to Khmer's low-resource status, mixed-script usage, named entities, and domain-specific vocabulary. At the same time, the strong hybrid performance shows that dense retrieval is not useless: it provides complementary signals, although the lexical signal remains dominant.
\begin{figure}[t] 
\centering \includegraphics[width=\textwidth]{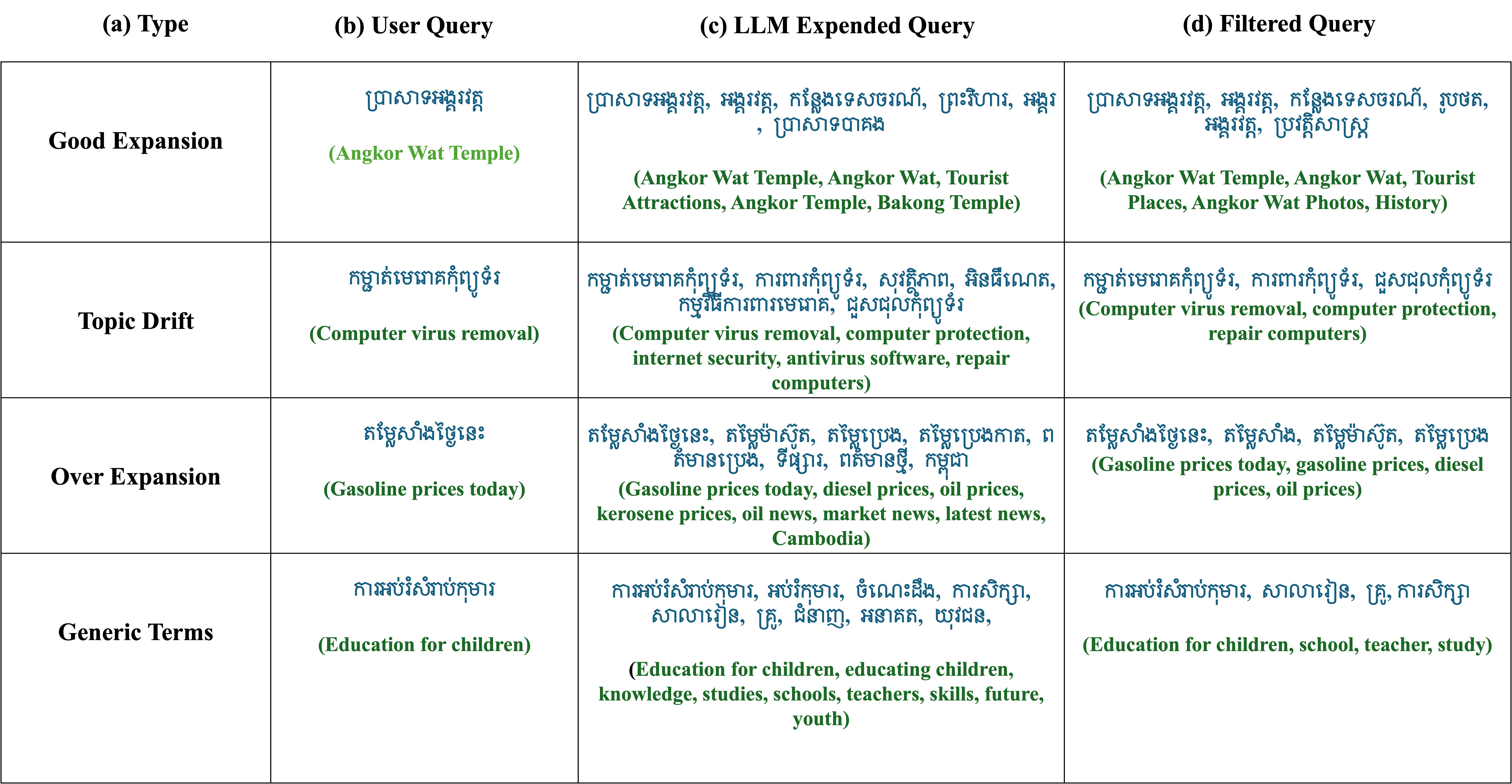} 
\caption{Qualitative examples of LLM-assisted query expansion behavior. While some expansions preserve the original intent and add useful related terms, other cases introduce topic drift, generic terms, or over-expanded reformulations. These examples illustrate why direct LLM expansion does not consistently improve Khmer retrieval performance.} 
\label{fig:llm_behavior} \end{figure}
\subsection{Impact of LLM-Assisted Query Expansion}
\label{subsec:llm_expansion_results}

LLM-assisted query expansion does not outperform non-expanded retrieval. For example, BM25 with Qwen2.5-0.5B expansion achieves 0.824 Recall@10 and 0.770 nDCG@10, which is lower than the original BM25 baseline. Similarly, hybrid retrieval with Qwen2.5-0.5B expansion reaches 0.788 Recall@10 and 0.729 nDCG@10, below the non-expanded hybrid baseline.

The same pattern holds for Qwen2.5-3B. Hybrid retrieval with Qwen2.5-3B expansion achieves 0.868 Recall@10 and 0.788 nDCG@10. This is better than hybrid retrieval with Qwen2.5-0.5B expansion, but still below the non-expanded hybrid baseline. These results suggest that direct LLM query expansion may dilute the original lexical signal or introduce terms that shift the query away from the most relevant documents.

\subsection{Effect of LLM Size}
\label{subsec:llm_size}

Although LLM expansion does not surpass the non-expanded baselines, model size clearly affects expansion quality. Qwen2.5-3B consistently outperforms Qwen2.5-0.5B under both dense and hybrid retrieval settings. For dense retrieval, Qwen2.5-3B improves Recall@10 from 0.365 to 0.440 and nDCG@10 from 0.318 to 0.372 compared with Qwen2.5-0.5B. For hybrid retrieval, Qwen2.5-3B improves Recall@10 from 0.788 to 0.868 and nDCG@10 from 0.729 to 0.788.

This indicates that larger instruction-tuned LLMs generate more retrieval-useful query expansions than smaller models. However, the improvement is not enough to outperform the original-query BM25 or hybrid baselines. Therefore, LLM size helps, but direct expansion remains insufficient for robust Khmer retrieval.

\subsection{Effect of Filtering Expanded Queries}
\label{subsec:filtering_results}

The filtered expansion variants generally perform worse than the raw expanded-query variants. For example, Hybrid + Qwen2.5-3B decreases from 0.868 Recall@10 and 0.788 nDCG@10 to 0.838 Recall@10 and 0.755 nDCG@10 after filtering. A similar decrease is observed for Qwen2.5-0.5B. This suggests that simple filtering removes not only noisy terms but also useful semantic cues.

These results show that filtering LLM expansions is not straightforward. Generic stopword-like filtering may reduce topic drift in some cases, but it can also weaken useful reformulations. More advanced filtering strategies should preserve named entities, maintain query intent, and adapt to different query types.

\subsection{Qualitative Analysis of LLM Expansion}
\label{subsec:qualitative_analysis}

Figure~\ref{fig:llm_behavior} illustrates common behaviors of LLM-assisted query expansion. Some expansions preserve the original query intent and add useful related terms. However, other cases introduce topic drift, generic terms, or overly broad reformulations. For example, a specific entity query may be expanded into a general tourism or culture query, causing the retriever to return broad documents rather than the target document. In other cases, the LLM adds general terms such as information, history, document, or Cambodia, which may match many documents and reduce ranking precision.

These qualitative examples help explain the quantitative results. LLM expansion can be helpful when it adds precise related terms, but it can hurt retrieval when it weakens the original lexical signal or changes the query focus. This is particularly important for Khmer, where exact entity matching and character-level overlap remain strong retrieval signals.

\subsection{Discussion, Limitations, and Future Work}
\label{subsec:discussion_limitations}

The results lead to three main observations. First, character n-gram BM25 is a strong baseline for Khmer web retrieval. Second, hybrid retrieval performs close to BM25, suggesting that dense semantic representations provide complementary but limited signals. Third, LLM-assisted query expansion depends strongly on model size and expansion quality. Larger LLMs produce better expansions than smaller LLMs, but direct expansion still does not outperform strong non-expanded retrieval baselines. These findings do not suggest that LLMs are ineffective for Khmer retrieval. Rather, they show that unrestricted query expansion alone is insufficient. Future LLM-assisted retrieval should consider query-type-aware prompting, entity-preserving expansion, multi-query fusion, retrieval-aware reranking, and stronger human-verified evaluation. The relevance labels are silver labels constructed from source-document matching, BM25 candidate retrieval, rule-based scoring, category agreement, and partial human verification, which may favor lexical retrieval. Finally, dense retrieval is evaluated with multilingual-E5-small, and LLM-assisted retrieval uses direct expansion with simple filtering. Future work will extend KSE-Web to PDF and OCR-based documents, collect more natural user queries, add larger human-verified relevance judgments, and evaluate stronger multilingual, Khmer-adapted, and LLM-based retrieval methods.

\section{Conclusion}
\label{sec:conclusion}

This paper presented KSE-Web, an analysis of hybrid retrieval and LLM-assisted query expansion for low-resource Khmer semantic search using a Khmer web retrieval dataset. We constructed the dataset from approximately 17K candidate Khmer web titles and retained 3K cleaned full-text Khmer documents after filtering, normalization, deduplication, and document-length control. The dataset includes 300 manually reviewed user-style Khmer queries and 5,412 silver relevance labels with partial human verification. Our experiments compared character n-gram BM25, multilingual dense retrieval, hybrid BM25+dense retrieval, and LLM-assisted query expansion using Qwen2.5 models. The results show that character n-gram BM25 achieves the strongest overall performance, with 0.943 Recall@10 and 0.876 nDCG@10. Hybrid retrieval performs closely, reaching 0.929 Recall@10 and 0.871 nDCG@10, while dense retrieval alone performs substantially lower. LLM-assisted query expansion does not outperform non-expanded retrieval, although Qwen2.5-3B produces stronger expanded-query results than Qwen2.5-0.5B. These findings show that LLMs should not be assumed to improve low-resource retrieval automatically. For Khmer semantic search, character-level lexical retrieval remains a strong baseline, dense retrieval provides complementary but limited semantic signals, and LLM-assisted query expansion requires careful control to avoid topic drift and loss of lexical specificity. We hope this study provides a useful foundation for future research on Khmer retrieval, low-resource semantic search, and LLM-assisted document understanding.

\bibliographystyle{splncs04}
\bibliography{sn-bibliography}
\end{document}